\def\year{2020}\relax
\documentclass[letterpaper]{article} 
\usepackage{aaai20}  
\usepackage{times}  
\usepackage{helvet} 
\usepackage{courier}  
\usepackage[hyphens]{url}  
\usepackage{graphicx} 
\usepackage{graphicx}  

\usepackage[utf8]{inputenc} 
\usepackage{booktabs}       
\usepackage{amsfonts}       
\usepackage{nicefrac}       
\usepackage{microtype}      
\usepackage{subcaption}
\usepackage{mathrsfs}
\usepackage{algorithm}
\usepackage{algorithmic}

\title{Concept Drift Detection and Adaptation with Weak Supervision on Streaming Unlabeled Data}
\author{Abhijit Suprem}
 
\begin{document}

\maketitle

\begin{abstract}
  Concept drift in learning and classification occurs when the statistical properties of either the data features or target change over time; evidence of drift has appeared in search data, medical research, malware, web data, and video. Drift adaptation has not yet been addressed in high dimensional, noisy, low-context data such as streaming text, video, or images due to the unique challenges these domains present. We present a two-fold approach to deal with concept drift in these domains: a density-based clustering approach to deal with virtual concept drift (change in statistical properties of features) and a weak-supervision step to deal with real concept drift (change in statistical properties of target). Our density-based clustering avoids problems posed by the curse of dimensionality to create an evolving ‘map’ of the live data space, thereby addressing virtual drift in features. Our weak-supervision step leverages high-confidence labels (oracle or heuristic labels) to generate weighted training sets to generalize and update existing deep learners to adapt to changing decision boundaries (real drift) and create new deep learners for unseen regions of the data space. Our results show that our two-fold approach performs well with $>$90\% precision in 2018, four years after initial deployment in 2014, without any human intervention.
\end{abstract}

\section{Introduction}
Machine learning and its variants are widely used for a variety of tasks, from image object detection to text sentiment analysis, to event detection. More recently, each application domain has seen a shift from static machine learning on offline data-sets to the streaming mode where data is continuous, with research primary focused on space and time efficiency on training \cite{streamingclass2,streamingclass3}. In practice, however, these approaches do not address a key feature of the streaming mode that is not present in offline data: concept drift. 

The concept drift phenomenon is characterized by the continuous changes in the latent data generating function over time. We call data that exhibits such concept drift as evolving data, or evolving streams. There are several works on concept drift adaptation \cite{conc_drift_active_shan,conc_drift_almeida,conc_drift_costa,conc_drift_demello,conc_drift_windows}, most of which assume non-adversarial concept drift where: (a) streaming data has high quality, with little to no noise, (b) drift direction, type, and scale are known, (c) there is immediate and proportional feedback available to perform model correction, and (d) the streaming data exhibits strong-signal characteristics. These combinations of assumptions describe a closed dataset where data distribution parameters are known \textit{apriori} and extensive subject matter expertise is available to adjust distribution parameters and models as and when drift occurs.

We present a drift detection and adaptation approach for evolving data without making limiting, closed-dataset assumptions. Our work is influenced by works in drift detection and adaptation but is designed for adversarial drift conditions. We test our work in event detection on web data, specifically on weak-signal events where the relevant data is dwarfed by irrelevant data and noise. We select fast, accurate disaster detection from web data as a high-impact domain that has several adversarial drift conditions: (a) low quality of streaming data, since our dataset is composed of short-text streams \cite{short_text,short_text_sriram}; (b) unknown and unbounded drift, due in part to lexical diffusion \cite{lex_diff} and random shifts in user behavior; (c) absence of timely feedback due to scale of data - manually labeling of even 0.01\% of streaming web data ($>$500M samples per day) will require more than 20 workers each day to work continuously for 8 hours; and (d) weak-signal events with an abundance of irrelevant data and noise - our disaster dataset is an ongoing collection of live social and news feeds, and even with keyword search and filtering on disaster type, almost 94\% of data is drifting noise with time-varying characteristics that must be eliminated with fast-updating learning models.

\begin{figure*}[t]
	\centering
	\begin{subfigure}{0.35\textwidth}
		\centering
		\includegraphics[width=\textwidth]{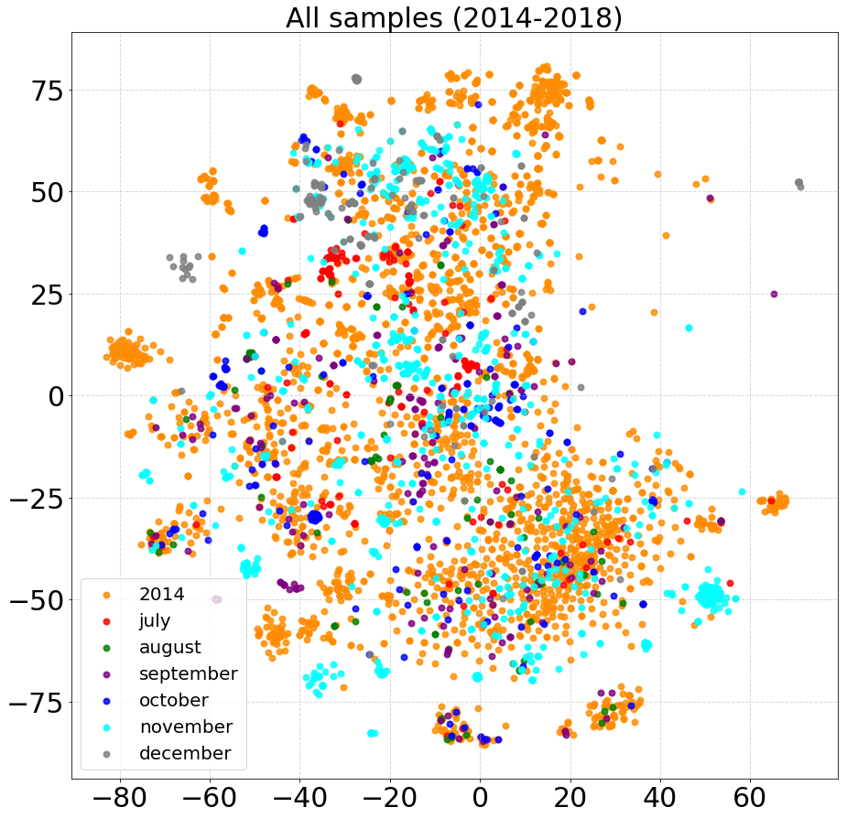}
		\caption{Drift in all samples (2014-2018)}
		\label{fig:drifta}
	\end{subfigure}%
	~\quad
	\begin{subfigure}{0.35\textwidth}
		\centering
		\includegraphics[width=\textwidth]{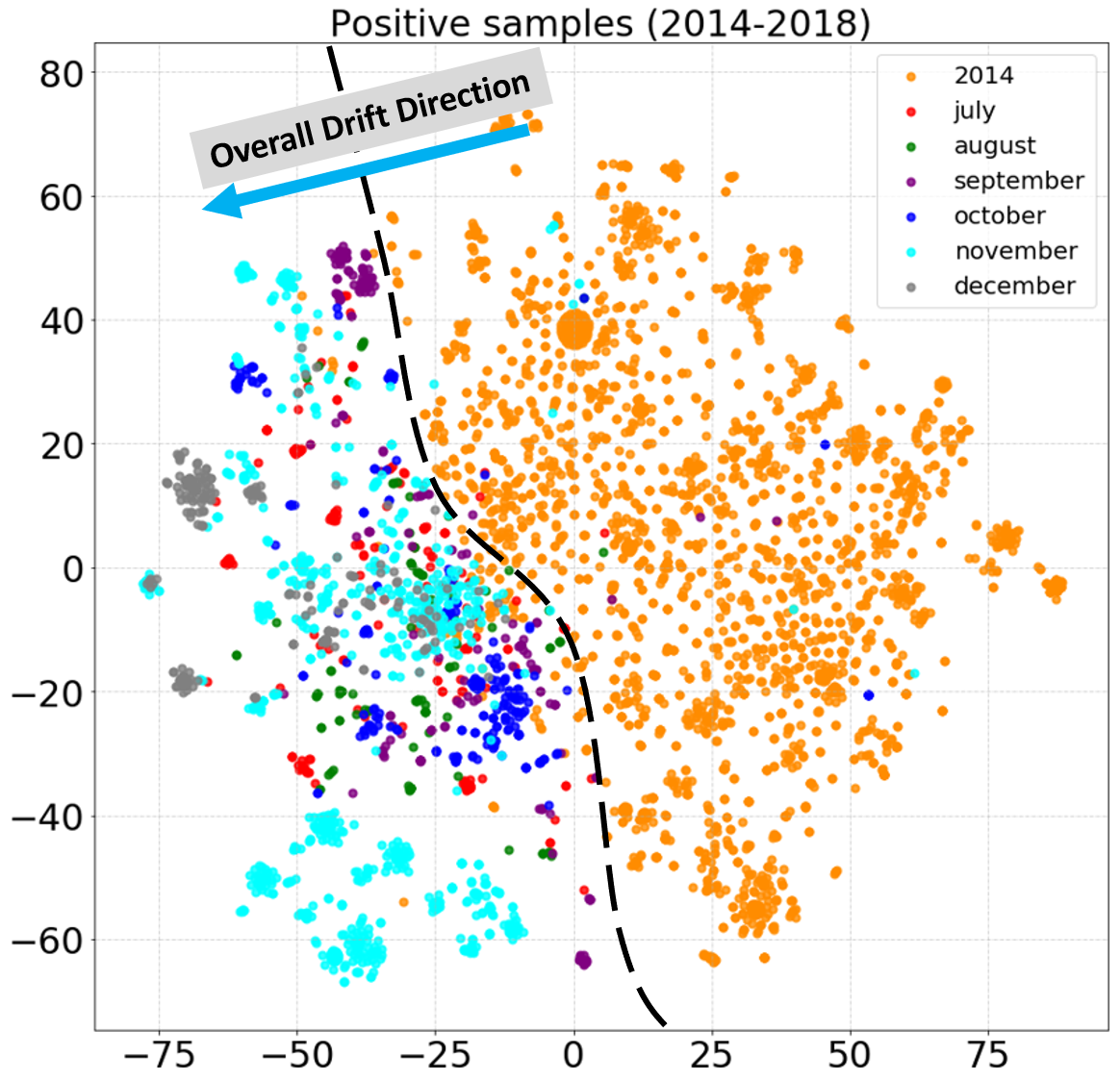}
		\caption{Virtual drift in positive samples (2014-2018)}
		\label{fig:driftb}
	\end{subfigure}
	~\quad
	\begin{subfigure}{0.35\textwidth}
		\centering
		\includegraphics[width=\textwidth]{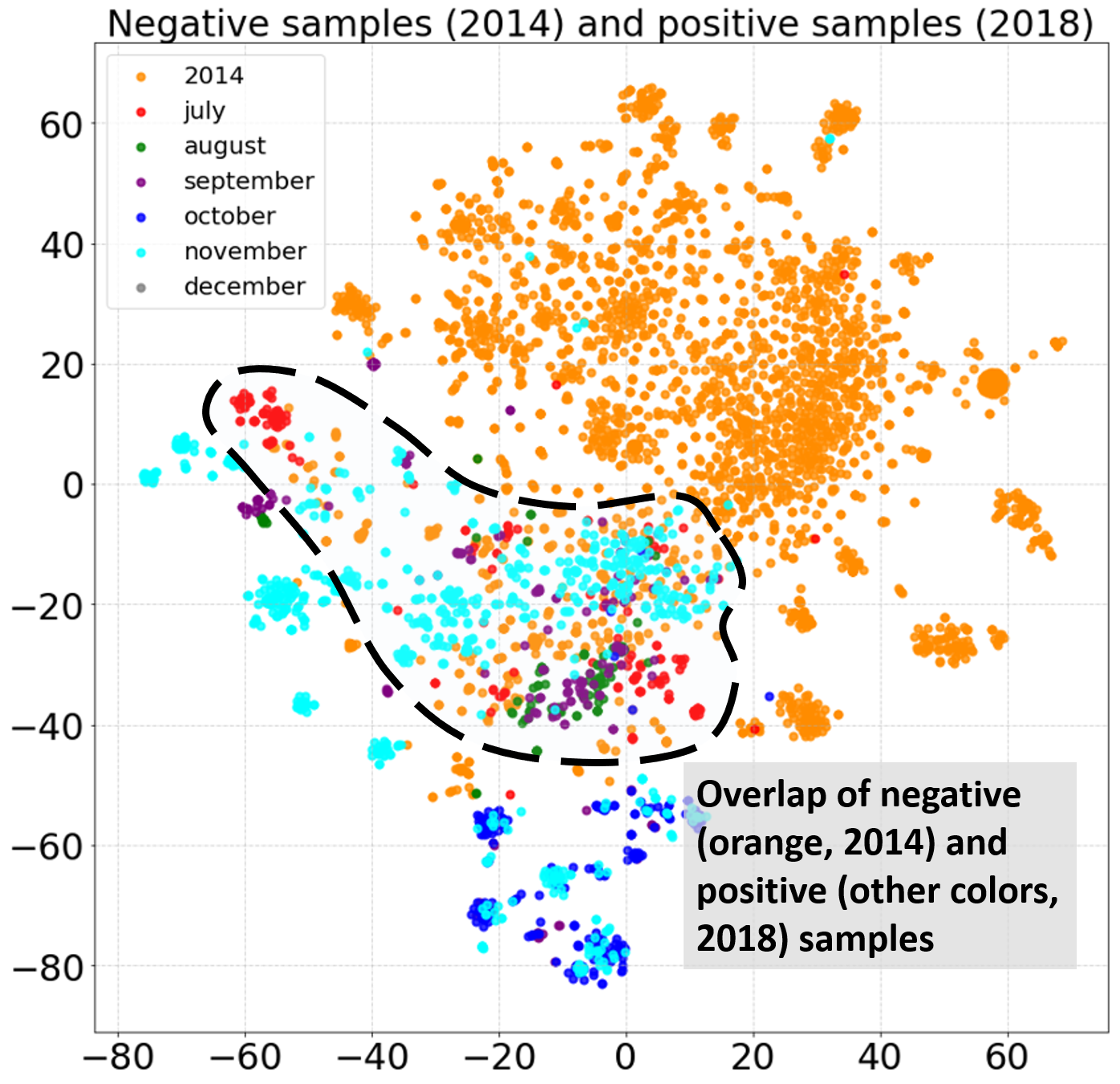}
		\caption{Real drift with positive samples (2014) compared with negative samples (2018)}
		\label{fig:driftc}
	\end{subfigure}
	~\quad
	\begin{subfigure}{0.35\textwidth}
		\centering
		\includegraphics[width=\textwidth]{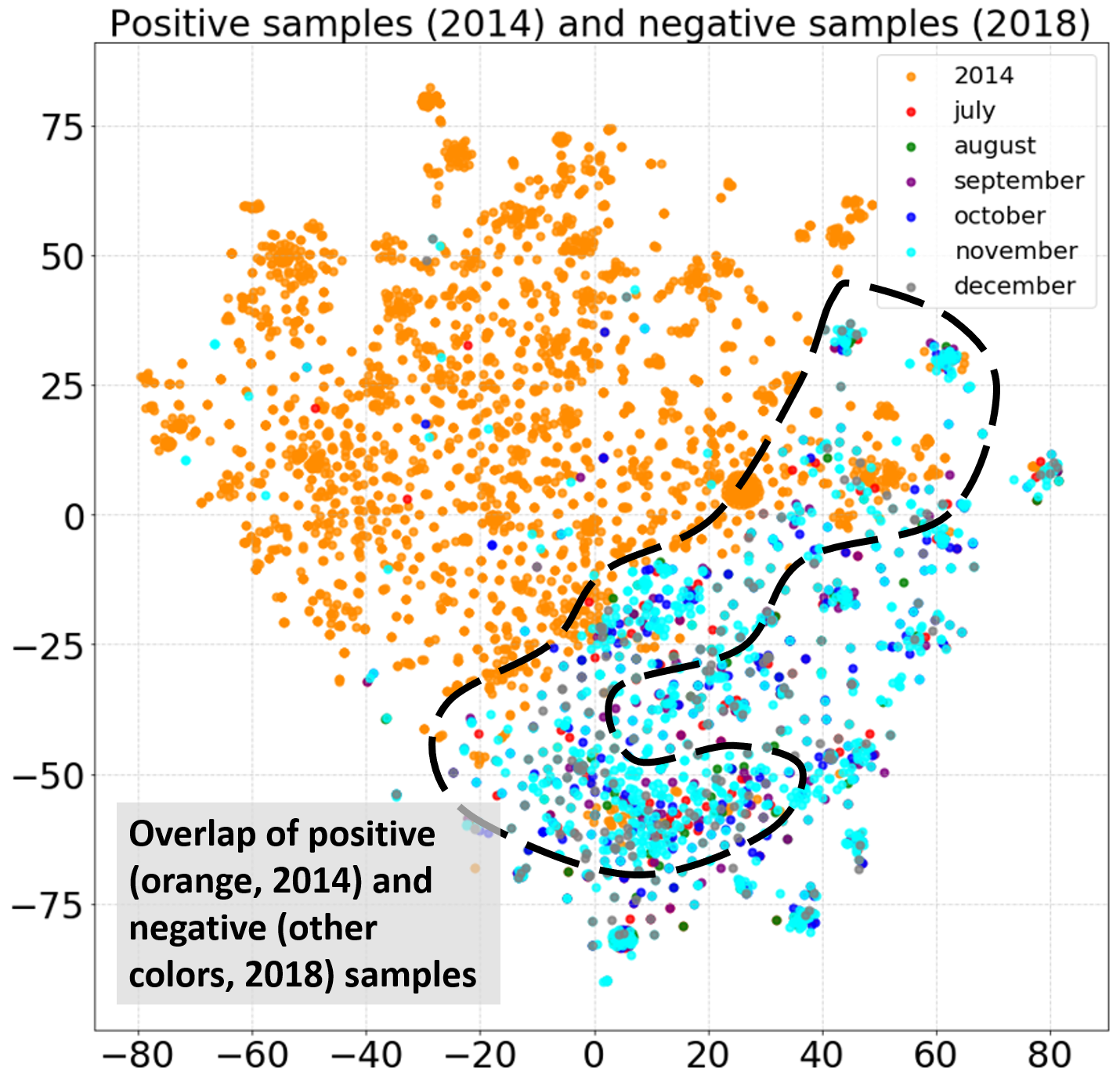}
		\caption{Real drift with negative samples (2014) compared with positive samples (2018)}
		\label{fig:driftd}
	\end{subfigure}
	\caption{Drift across multiple years in disaster detection dataset (axes correspond to raw tSNE scores)}
	\label{fig:driftevidence}
\end{figure*}

Our contributions are as follows: (1) we develop an implicit concept drift detection approach that does not rely on oracle feedback; our drift detection method is resistant to noise in our real-world data and is designed with curse of dimensionality in mind by tracking the high-density set of in-distribution samples; (2) we present a weak-supervision method to significantly augment our training data during inter-window update phases. Our approach generates weights for data without oracle labels with a novel derived confidence score that measures the agreement between non-oracle labeled samples and its closest oracle labeled neighbor.

\section{Preliminaries}
\subsection{Concept Drift}
We define concept drift in terms of events $\mathbf{E}$ and signals $\mathbf{S}$. Each data point $P_i$ is a distribution over events $P(E_a|P_i)$, $E_a\subset\mathbf{E}$, including a desired event $E_d\in E_a$. $P_i$ is also defined from a generative model $P(P_i|\mathbf{S})$, so $$E_a=\sum_i^k a_i S_i$$ where each $S_i$ is a concept of the data. Under this formulation, concept drift occurs when the distribution of $a_i$ changes, usually over time. However, decomposing $P_i\in\mathbb{R}^n$ into its exact component generative signals $\{S_k\}^i$ is difficult, since each concept lies in a unstructured dark knowledge space \cite{darkspace} and the mapping from signals to data points $f:\Omega_S\rightarrow\Omega_P$ requires large amounts of data to build a complete mapping (e.g. w2v \cite{w2v} maps words to a semantic vector space using a well-defined, large text corpus). Unfortunately, streaming and natural data change too quickly or are too sparse \cite{lex_diff} for such mappings.

Concept drift falls under two categories: \textit{real} concept drift, where the decision boundaries between classes change, and \textit{virtual} concept drift, where the distribution of $a_i$ changes without any change in boundaries. The former can be addressed with model updates using newly labeled data, while the latter can be addressed with models that have strong generalization capabilities. A combination of \textit{real} and \textit{virtual} drift poses challenges, however: model update fails if there is a lack of new labeled data (the case with weak-signal events) and generalization fails when the decision boundary itself shifts. 

Literature has identified several variants of these drift: \textit{gradual} drift causes slight changes to decision boundary, in contrast to \textit{sudden} drift; \textit{cyclic} drift causes signals that have disappeared ($a_i=0$) to reappear; cyclic drift may be \textit{periodic}, such as seasonal concepts (e.g. elections) or \textit{aperiodic}; \textit{flash} drift represents ephemeral concepts (flash drift is not noise, since it causes spike in a single $a_i$, while noise applies a bias to all $a_i$).  

\subsection{Evidence of drift in our data}
\label{sec:data}
We show concept drift in our disaster detection dataset spanning from 2014 through 2018. Each data point is a social media post, filtered on disaster-related keywords (e.g. \textit{landslide}, \textit{mudslide}, and \textit{rockslide} for landslides while filtering out unrelated keywords); this initial keyword filtering still leaves $\sim$94\% noise. We use a combination of long-term heuristics and manual filters to remove irrelevant samples. Data points remaining after this second filtering step require learning models to classify as relevant or irrelevant. We encode each text data point as a vector, and dimensionally reduce with t-SNE to better represent the pairwise distance between points in Figure~\ref{fig:driftevidence}.

Figure~\ref{fig:drifta} shows embedding of both positive and negative samples from 2014 through 2018. While there are some samples in 2018 that are outside the bounds defined by the 2014 data, drift is not conclusive. It becomes apparent in Figure~\ref{fig:driftb}, where positive samples in 2018 are clearly separated from positive samples in 2014. So, classifiers trained to identify relevant samples may have increased false negatives if they do not adapt to the drift in 2018. This is an example of \textit{virtual} drift, as it is not yet clear if the decision boundary itself has shifted. Figure~\ref{fig:driftc} presents an example of \textit{real} drift, where some samples considered negative, or irrelevant to the event in question, are indistinguishable from positive samples in 2018. This shift in classes themselves represents the \textit{real} drift – classifiers trained to recognize irrelevant samples in 2014 would make increase false negative errors in 2018. Finally, we show \textit{real} drift in the reverse direction in Figure~\ref{fig:driftd} – there are positive samples in 2014 that are closer to negative samples in 2018, increasing chance of false positives.

\section{Related Work}
\label{sec:related}
Recent works have focused on adapting the static classifiers to the dynamic, streaming domain~\cite{conc_drift_almeida,adversarial_drift,conc_drift_demello,conc_drift_active_shan,rejection}. However, these approaches keep many of the data assumptions of the static models; a thorough survey is available in \cite{gama_drift_a}. We briefly describe some of the assumptions. The \textbf{closed data} assumption is where streaming data is well specified by the training data and in case of drift, large amounts of ground truth labels can be quickly generated. However, naturally drifting data distributions such as language are difficult to characterize and label~\cite{twt_lang_model}. Related is the \textbf{immediate feedback} assumption, where oracle labels are available quickly for drift updates. Drift adaptive approaches often use a single labeled dataset as a good model of all real-world data ~\cite{gama_drift_a,gama_drift_b}. As such, many current works use closed and/or synthetic data where the type, scope, and size of drift is known~\cite{conc_drift_almeida,rejection,paired_learner,gama_drift_c}. The \textit{Knowledge Maximized Ensemble} approach \cite{kme} attempts knowledge-agnosticy by combining multiple drift detectors and update procedures to remain adaptive to multiple drift types. The approach in \cite{streamingclass1} tracks real drift and maintains memory of irrelevant data to retroactively change existing data's classes for future training. It does not handle virtual drift in the data space, however.

\section{Approach}
We first give an overview of our event detection pipeline to place our drift detection and weak-supervision labeling approach in context. Drift in web data is continuous and unpredictable. Further, the scale of web data makes oracle feedback impossible. We use the knowledge transfer approach from \cite{assed}, specifically the Heterogenous Data Integration process, to map data points from noisy, but abundant sources (social media, web data) to reputable, but scarce ground-truth events (news articles, government reports of disasters, etc.) to generate high confidence labels. Knowledge transfer can label $\sim$5\% of streaming data. The rest passes through our classification step. Additionally, knowledge transfer is only able to label positive samples. Data points that could not be mapped cannot be considered negative samples, since they could be data points in locations without reputable source coverage (e.g. disaster not covered by news or tracking agencies).

\subsection{Unlabeled Drift Detection}
\label{sec:ddm}
There are several works on out-of-distribution detection \cite{oodd1,oodd2,oodd3,oodd4} using novelty detectors, adversarial autoencoders, and generative adversarial networks. However, such approaches are not suitable in our situation, where the majority of samples are noise. We also need to address virtual drift, where the distribution of both relevant and irrelevant points changes without changing the decision boundary itself. Autoencoder-based methods are effective when the majority class is relevant and the minority class is novelty. We need to capture the distribution of all classes in the current streaming window to track the virtual drift across labels. We also tested recently proposed model confidence methods \cite{md3a,md3b} that track sample density immediately adjacent to classifier margins. These approaches fail in high-dimensional drifting streams due to the curse of dimensionality: since the volume of the unit hypersphere approaches zero, most points are at the corners of the hypercube, making $L_2$ distance measures (i.e. Euclidean) ineffective, leading to models that are always confident.

Our drift detection approach uses the common Kullback-Leibler divergence test on two windows – the model window of data points the current models are trained on, and the streaming window of incoming data points. We compare the two distributions on a high-density band of points, with density determined with respect to a distance metric between data points and their centroid, or set mean of the points. 

\begin{figure*}[t]
	\begin{center}
		\includegraphics[width=0.7\textwidth]{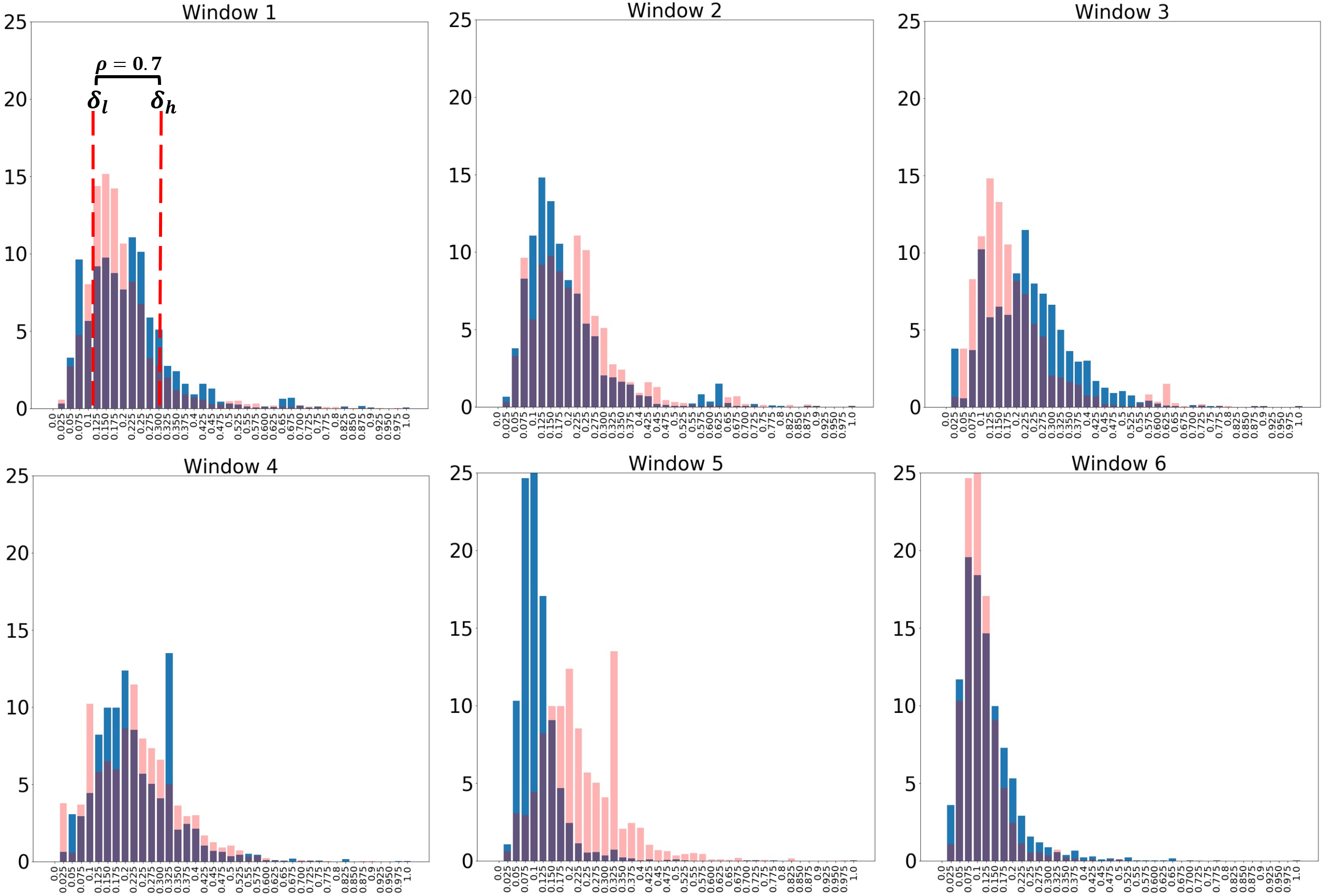}
	\end{center}
	\caption{$\rho$ distribution using distance metric as the divergence criteria; we show difference between windows Euclidean metric (previous window is light shaded, current window is dark shaded).}
	\label{fig:intuition}
\end{figure*}

We define this high-density band as follows: let $D_k$ be the dataset (or the set of points) in window $w_k$, with data points ${x_1,x_n,...,x_N}\in X \in\mathbb{R}^n$, where X is the data space of all points in the current window. The window $D_k$ has centroid 

$$D_k^C=N^{-1}\sum_N x_i$$ 

Let $f_D(x)$ be the continuous density function of $D$ estimated on a normalized distance metric $d:\mathbb{R}^n\rightarrow[0,1]$, where $d$ measures the distance between any point $x_i\in X$ and $D_k^C$. Then, the $\rho$-density band, with $\rho\in[0,1]$, is a region around the centroid that contains $\rho$ probability mass of the data window. We consider this as a banded region $[\delta_l, \delta_h]$, where $0\leq\delta_l < \delta_h\leq 1$, and calculate the region bounds as:

\begin{equation}
\int_{\delta_l}^{\delta_h}f_D(x)dx=\rho
\end{equation}

Our intuition for obtaining the bounds to set $\rho = [\delta_l, \delta_h]$ is directly related to the curse of dimensionality; for any set of points in high-dimensional space, the volume of the unit hypersphere tends to zero\footnote{$V(d)=0.5^d\pi^{0.5d}/\Gamma(0.5d + 1)$}; consequently, most points occur near the corners relative to the centroid. The $\rho$-band is then a density band (or equivalently, if $\rho>0.5$, a high-density band, since the band would contain greater than half of the data window's points) around the centroid, where the hyperspherical region of radius $\delta_l$ around the centroid is mostly empty. Additionally, this sparsely populated hypersphere region of a dataset may itself be part of the $\rho$-band of a separate dataset in the $X$ data space.

Our intuition is borne out in Figure~\ref{fig:intuition}, which shows a histogram of the distances of points from the centroid in several data windows. Each point is a post from the disaster detection dataset as described in \textbf{Preliminaries} section. We find that the hypersphere region is empty, and most points are concentrated in a band around the centroid, with a long-tailed distribution of points. We also show the band for $\rho=0.7$ in Window 1 of the figure, where the bounds contain 70\% of the data points. The $\rho$-band of our data is approximated by $\mathscr{N}(\mu,\sigma^2)$, where $\mu, \sigma$ can be estimated.

Then, given the $\rho$-band of a model window $w_M$, we can define the $\rho$-band of the current data stream $w_S$ with respect to $w_M$ as $[\delta_l, \delta_h]_{w_M}$ initially, and update it with the new data points from $w_S$. Then, drift detection between the classifier window(s) and the streaming window is performed with a modified Kullback-Leibler divergence on the $\rho$-band with the given distance metric (e.g. $d(x_i,D_k^C)$), since we are comparing the partial probability distributions contained within the $\rho$-band.

Let $$D_{KL}(P_A||P_B)=-\sum_{x_i\in X}P_A(x_i)\log(P_B(x_i)/P_A(x_i))$$ be the standard KL metric on two distributions modeling a new data point $x_i$ - the prior $P_A$ and the posterior $P_B$, where $w_M$ is the prior and $w_S$ is the posterior. Then, let $x'=d(x,C_A)$, where $d$ is a distance metric and $C_A$ is the centroid of A (or $w_M$). We obtain $x'_A$ and $x'_B$ from the prior and live distributions of $w_M$ and $w_S$, respectively. Since KL is undefined if $P_A(x')=0$, we make the approximation $P_A(x')=\epsilon$ if $P_A(x')=0$, where $\epsilon=\min(P(x'))$ ($\epsilon$ is necessarily part of the $\bar{\rho}$-density band $[0,\delta_l)$ or $(\delta_h,1]$). We allow a smoothing period between windows to incorporate the new stream if drift is detected; our divergence metric has two hyperparameters: the smoothing window $w_L$ and the KL threshold $\theta_{KL}$. 

\begin{figure*}
	\begin{center}
		\includegraphics[width=0.7\textwidth]{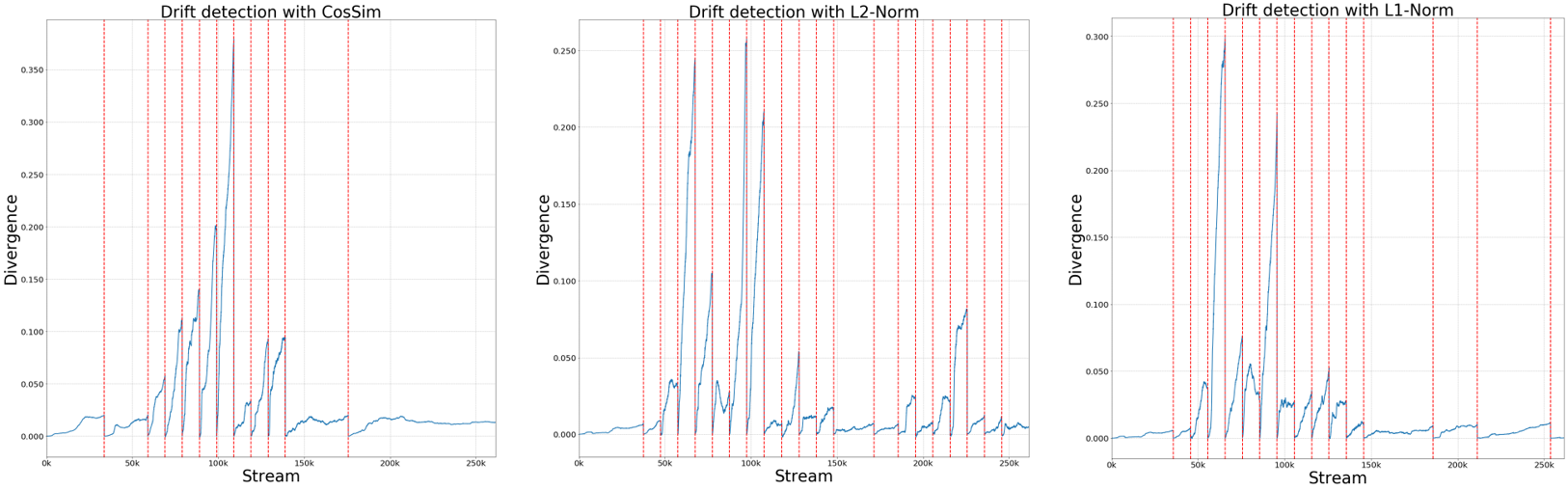}
	\end{center}
	\caption{Drift detection under Cosine Similarity, L2-Norm, and L1-Norm distance metrics using divergence of $\rho$-band distributions. Each vertical red line is a instance of virtual drift detected by the unsupervised drift detection algorithm}
	\label{fig:kldetection}
\end{figure*}

We test drift detection under several distance metrics and show results for Cosine Similarity, L2-Norm, and L1-Norm in Figure~\ref{fig:kldetection}. Both Cosine Similarity metric and L1-Norm identify virtual drift early in the stream. As more of the data-space is discovered and added to ensembles through model and general memories, instances of drift detection decrease. However, the L2-Norm is far more susceptible to noise and continues to detect false-positive drift late in the stream, forcing unnecessary updates to models during the weak supervision step and leading to poorer performance. We cover this performance deterioration in Results.

\subsection{Drift Adaptation}
\label{sec:driftadaptation}
Drift detection is half of the solution; post detection, existing models need to be updated and new models generated for drifted data. We now describe our \textit{virtual} and \textit{real} drift adaptation approaches, where our \textit{real} drift adaptation uses weak supervision weak supervision to augment the oracle-labeled data (only $\sim$5\% of the streaming data can be labeled using knowledge transfer; we label the remaining 95\% with weak supervision).

A classifier $M$ is a mapping $f_M:\mathscr{M}\rightarrow\mathscr{Y}$ to class labels $y'\in\mathscr{Y}$, where $\mathscr{M}$, the training and testing data of model $M$, specifies a region in the data space $\Omega_D\in\mathbb{R}^n$. The traditional, offline mode has typically characterized $\mathscr{M}$ as representative of the universe of data points (see \textbf{Related Work} section); in fact, this is a core criterion of generalizability. Models are designed to generalize with the assumption that training data is highly representative of unseen data. 

This assumption is unsuited for streaming drifting data where the training data in one window is dissimilar to training data in another window, and any window contains only a subset of all data. A key aspect of unpredictable drift is that there are signals that may not have been encountered yet. The target of a single generalizable model or ensemble of models fails when the decision boundaries themselves shift. Our goal, then, is to build and continuously update a temporally evolving collection of mappings $f_{\{M\}^k}:\{\mathscr{M}\}^k\rightarrow\mathscr{Y}$ over $\Omega_D$, where each $\mathscr{M}_i$ specifies a window in $(\mathbb{R}^n, w_i)$, and $w_i$ is a streaming dataset window at time $t_i$. Windows are separated by the unlabeled/unsupervised KL-based drift detection mechanism in \textbf{Unlabeled Drift Detection} section. 

Adapting to \textit{virtual} drift involves generating new mapping $f_{M_n}$ for points in $\mathscr{M}_n$, where $\mathscr{M}_n\cap\{M\}^k=\emptyset$. Adapting to \textit{real} drift involves updating existing mappings $f_{M_i}$, $M_i\subset \{M\}^k$ to reflect updated decision boundaries on the samples in $\mathscr{M}_i$.  We propose the following algorithms to adapt to \textit{virtual} and \textit{real} concept drift simultaneously.

\subsection{Adapting to Virtual Drift}
We first describe our \textit{virtual} drift adaptation algorithm. Let $\{M\}^k$ be the set of models over $\{\mathscr{M}\}^k\in\mathbb{R}^n$ at current streaming window  $w_s$, and $x_i\in\Omega_D$ be a new point from $X$. We define an ensemble selection policy as a hyperparameter that defines how ensembles should be selected. Some examples of an ensemble selection policy include: set of all \textit{recent} models (where \textit{recent} indicates models created in the prior drift detection update step); high performing models over the entire set of models; high performing \textit{recent} models; $k$-nearest models based on distance between data point and centroids of the model's data window; or nearest $\rho$-band models where only models whose $\rho$-band contains $x_i$ are considered. 

Let $S_m:X\rightarrow \{M\}^m$ be any ensemble selection policy that selects the $m$-best models to classify $x_i$. We define two types of data memories - a model-specific memory $\mathscr{D}_{M_k}$ for each model $M_k$, and a single general memory $\mathscr{D}_G$. Then, for each $x_i$ and its ensemble $\{M\}^m\subseteq S_m (x_i)$, place $x_i$ in $\mathscr{D}_{M_k}$ if $x_i\in[\delta_l,\delta_h]_{M_k}$. Add $x_i$ to $\mathscr{D}_G$ if it is in no $\rho$-band. We also allow each model to generalize beyond the $\rho$-band by putting all data points from $[\delta_h,\lambda]$ in $\mathscr{D}_{M_k}$ in addition to adding it to $\mathscr{D}_G$ (where $\lambda$ is the generalization distance, $\delta_h < \lambda \leq 1$). The $\rho$-band is recomputed after updating $\mathscr{D}_{M_k}$ and $M_k$. $\mathscr{D}_{M_k}$ is used for \textit{real} drift adaptation. The general memory $\mathscr{D}_G$ stores recent discoveries to the known data space, allowing \textit{virtual} drift adaptation.

\subsection{Weak Supervision for Real Drift Adaptation}

\begin{figure*}
	\begin{center}
		\includegraphics[width=0.7\textwidth]{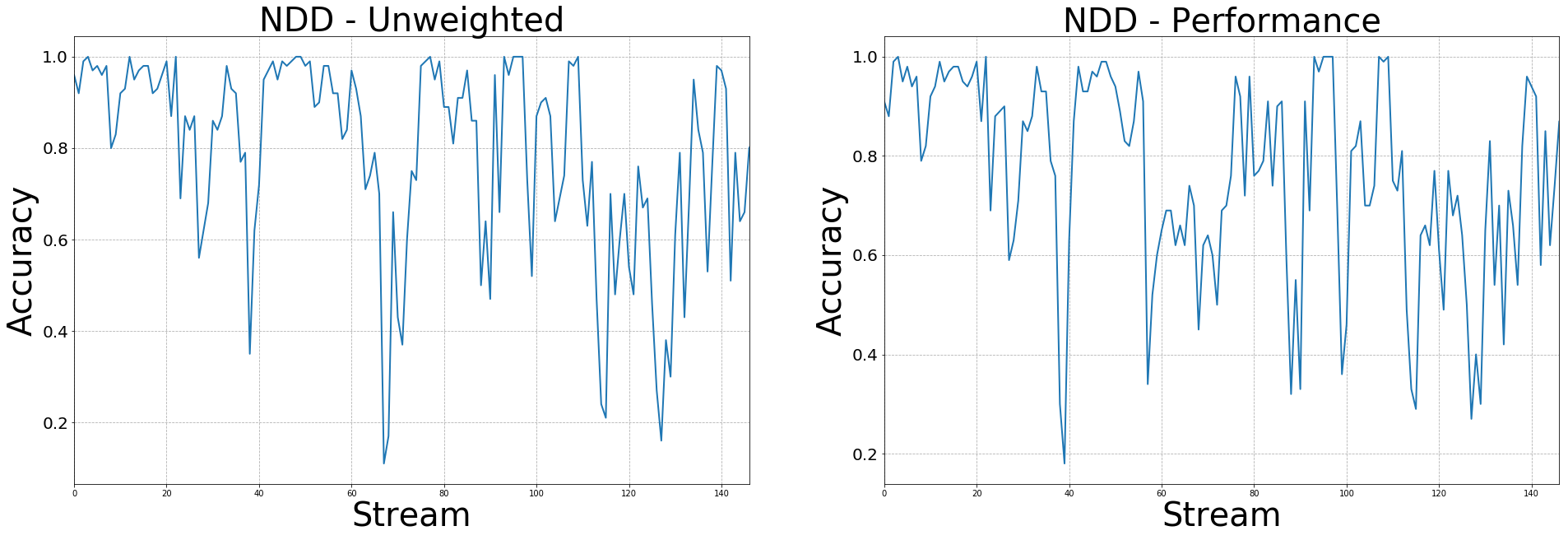}
	\end{center}
	\caption{Baseline performance with no drift detection (NDD). An ensemble of classifiers (neural networks, deep networks) is created at the initial training period. Throughout the data stream (from 2014 through 2018), the original models are used with either unweighted voted majority (left) or weights assigned based on performance. Each point is performance over 100 samples.}
	\label{fig:noupdates}
\end{figure*}

When drift is detected, we address \textit{real} drift for each model $M_k$ by using the data in $\mathscr{D}_{M_k}$.  As mentioned, \textit{virtual} drift is addressed by generating a new model $M_n$ on the data in $\mathscr{D}_G$. Using only oracle-labeled data (obtained from e.g. knowledge transfer, or HDI process from \cite{assed}) does not yield enough data for training and update. We augment oracle-labeled data with weak-supervision in each memory $\mathscr{D}_i$ as follows: for each non-oracle labeled $x_i\in \mathscr{D}_i$, we find the closest oracle-labeled $x_o$ with the same oracle label as $x_i$, and record the distance $d_w$ between $x_i$, $x_o$ using the same distance metric used for drift detection.

During training, each $x_o$ is weighted with a value of 1. Each $x_i$ is weighted on its distance from its closest agreement $x_o$ as $$w_i=\exp(\alpha\cdot d_w)$$

where $\alpha=-\ln(\theta_w/\lambda)$. $\theta_w$ is a hyperparameter for the desired weight (usually a small $\epsilon$) at a distance $\lambda$ from $x_o$. $\lambda$ is the generalization parameter from \textit{virtual} drift detection. We define it similar to $\rho$ to be a density range for generalization; to reduce number of parameters, we let $\lambda=\delta_h-\delta_l$ as the distance threshold to enforce $\theta_w$, allowing model $M_i$ to generalize within a $\rho$-band beyond $\delta_h$ of $\mathscr{D}_{M_i}$.

In our implementation, described in Algorithm 2, we build a KD-Tree on the oracle labeled samples for fast retrieval during the weak-supervision step.

\begin{algorithm}
	\caption{Adapting to \textit{Virtual} Drift}
	\label{alg:virtualdrift}
	\begin{algorithmic}
		\STATE $\mathtt{Parameters}$: $d$ (the distance metric, e.g. $\mathtt{CosSim}$, $\mathtt{L2}$, $\mathtt{L1}$); Ensemble model selection policy $S_m$; $\lambda$ (generalization distance)
		\STATE $\mathtt{Inputs}$: Current models $\{M\}_k$, $x_i$
		\STATE $\{M\}^m = S_m(x_i)$.
		\STATE $mem\_xi = \mathtt{False}$
		\FOR{$M_j \in\{M\}^m$}
		\STATE \# $D^j$ is the data set for model $M_j$, with $\rho$-band $[\delta_h^j, \delta_h^j]$
		\STATE $d'_{x_i} = d(x_i, D_K^j)$
		\IF {$\delta_l^j < d'_{x_i} < \delta_h^j$}
		\STATE $\mathscr{D}_{M_j} = \mathscr{D}_{M_j} \cup x_i$
		\STATE $mem\_xi = \mathtt{True}$
		\ENDIF
		\IF {$\delta_h^j \geq d'_{x_i} < \lambda$}
		\STATE $\mathscr{D}_{M_j} = \mathscr{D}_{M_j} \cup x_i$
		\ENDIF
		\ENDFOR
		\IF {\NOT $mem\_xi$}
		\STATE $D_G = D_G\cup x_i$
		\ENDIF		
	\end{algorithmic}
\end{algorithm}

\begin{algorithm}
	\caption{Adapting to \textit{Real} Drift}
	\label{alg:realdrift}
	\begin{algorithmic}
		\STATE $\mathtt{Parameters}$: $\lambda$ (generalization distance); $\theta_w$ (desired weight at $\lambda$)
		\STATE $\mathtt{Inputs}$: Models $\{M\}_k$; Memories $\mathscr{D}_G$, $\{\mathscr{D}_M\}^k$
		\FOR{$\mathscr{D}_{j}\in\{\mathscr{D}_M\}^k \cup \mathscr{D}_G$}
		\STATE $x_o$ is set of oracle-labeled points in $\mathscr{D}_j$, with label $y_o$
		\STATE $\mathtt{KT}=\mathtt{Build\_KDTree}(x_o)$
		\FOR{$x_i\in \mathscr{D}_{j} - x_o$}
		\STATE $x'_o, y'_o=\mathtt{KT}(x_i)$
		\STATE $w_i=\exp(-\ln(\theta_w/\lambda)\cdot d(x_i, x'_o)) \cdot (\mathbb{I}(y_i == y'_o))$
		\ENDFOR
		\STATE Update $M_k$ with $w$-weighted $\mathscr{D}_j$
		\ENDFOR
	\end{algorithmic}
\end{algorithm}

\section{Results}

We compare our unlabeled/unsupervised drift detection and adaptation approach against baseline performance without any drift detection or adaptation, and against explicit drift detection methods without weak supervision. Our experimental setup is as follows: our data-set is a live short-text stream that is encoded using $w2v$ to $\mathbb{R}^{300}$. We have shown drift in this dataset in Figure~\ref{fig:driftevidence}. The short-text stream is obtained from web blogs, Twitter, Facebook, and other web sources. We perform initial keyword filtering on relevant keywords (e.g. for landslide disaster detection, we use \textit{landslide}, \textit{mudslide}, and \textit{rockslide}, while performing initial filtering with \textit{election}, \textit{vote}, etc to remove samples clearly irrelevant to landslide disasters). We use a variety of text-classification networks \cite{fasttext,convclass,convclass2}
to classify data points from a live data stream from July 2014 through December 2018.

\subsection{Evaluation}


\begin{figure*}[t]
	\begin{center}
		\includegraphics[width=0.98\textwidth]{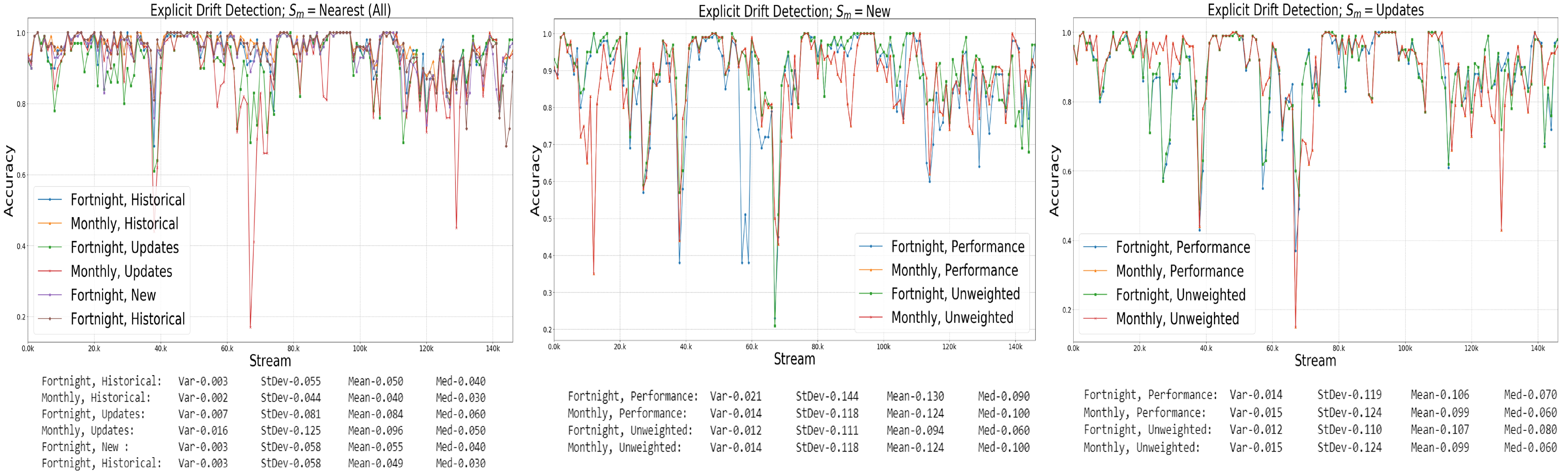}
	\end{center}
	\caption{Explicit drift detection + adaptation under 3 modes: \textit{Nearest}: Select \textit{top-k} models using NN search using $d(x_i,D_k^C)$; \textit{Updates}: Use only updated models; and \textit{Newest}: Use only newly generated models from $D_G$.}
	\label{fig:explicit}
\end{figure*}

\begin{figure*}
	\begin{center}
		\includegraphics[width=0.9\textwidth]{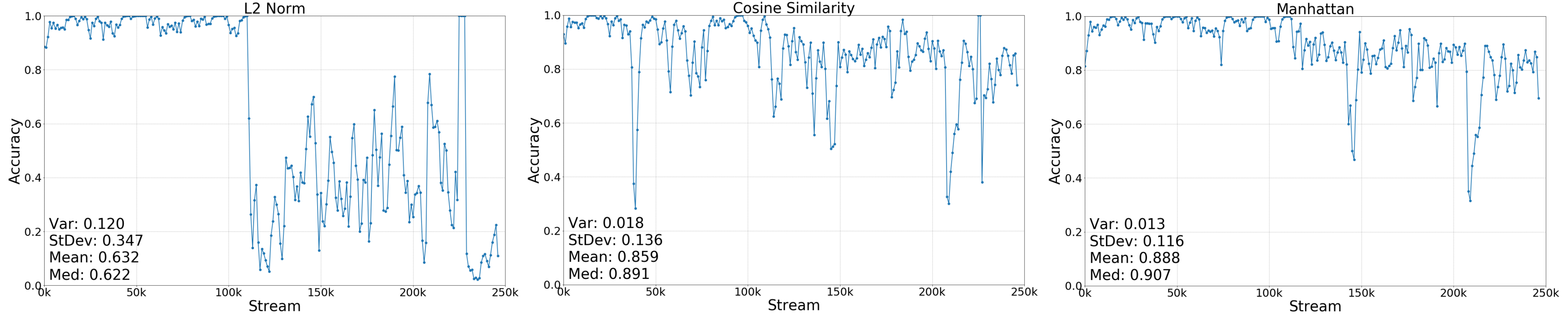}
	\end{center}
	\caption{Drift detection and adaptation with weak supervision under L2-Norm, Cosine Similarity, and L1 (Manhattan) Norm.}
	\label{fig:goodresults}
\end{figure*}




Baseline performance is shown in Figure~\ref{fig:noupdates} under two ensemble weighting schemes: unweighted and performance. \textit{Unweighted} uses unweighted average of all predictions. \textit{Performance} uses weighted ensemble, with each weight derived from model performance on its training window using $w_{M_k^t} = f_{M_k}\sum_a^n f_{M_k}$ where $f(M_k)$ is the performance of model $M_k$. Initial performance is high, with accuracy (\textit{f-score} measure) between 95-100\%. However continuous drift causes significant spikes in performance in both modes, with high variability in performance and lowest accuracy of 4\%.

Figure~\ref{fig:explicit} shows performance under explicit drift detection and update without weak supervision. Drift detection is conducted with DDM, EDDM, Page-Hinkley Test, and other explicit methods from \cite{gama_drift_a}. \textit{Top-k nearest models} selection approach (\textit{nearest} in Figure~\ref{fig:explicit}) sees variable performance up to mid-stream, when enough of the data space is discovered for more stable performance. However, increased virtual and real drift at the end of 2018 reduces average performance. Neither \textit{Updates} nor \textit{Newest} modes ($S_m$) show significant promise. Explicit drift detection relies on oracle labels and is not feasible in high data volume applications where feedback is expensive and slow. In each case, we perform the tests over half the stream as performance deteriorates too rapidly after 150k points.

Finally, we show performance with our unlabeled/unsupervised drift detection and updates with weak supervision in Figure~\ref{fig:goodresults}. For our unlabeled drift tracking with weak supervision, we test with various $\rho$ and show results for $\rho=0.5$. We approximate our $\rho$-band distribution as $\mathscr{N}(\mu,\sigma^2)$, based on the empirical  evidence in Figure~\ref{fig:intuition}. Since drift detection is unsupervised, we do not need to manually specify window sizes as in Figure~\ref{fig:explicit}. Drift windows in each case are determined by the detection thresholds from Figure~\ref{fig:kldetection}.

The $L_2$ norm performs poorly after significant \textit{real} and \textit{virtual} drift. The Manhattan distance metric performs better in the high dimensional case, with $>$90\% precision in 2018 (stream window $>$150k in Figure~\ref{fig:goodresults}). More interestingly, the unsupervised drift detection method is able to identify the \textit{virtual} drift using changes in $\rho$-band distribution. This \textit{virtual} drift corresponds to new regions of $\Omega_D$ appearing in $X$ that have not been encountered in any training data (tracked through $\mathscr{D}_G$). Our approach is able to both expand knowledge of $\Omega_D$ and adapt to changes in both existing and new regions of $\Omega_D$ discovered through the \textit{virtual} drift adaptation. Our results are obtained from fully automated drift detection and adaptation – at no point after initial training on 2014 data is human intervention used to adjust models, label samples, or select samples for model updates.

\section{Conclusions}
We present an approach for unsupervised drift detection and a  two-fold approach for simultaneous \textit{real} and \textit{virtual} drift adaptation in continuous drifting streams. Our drift detection approach builds an evolving \textit{view} of the data space by constructing and updating a series of mappings $f:\Omega_D\rightarrow\Omega_Y$, where $\Omega_D$ is the data space and $\Omega_Y$ is the label space. We adapt to \textit{virtual} drift in $\Omega_D$ by updating existing model spaces $\mathscr{M}\in\Omega_D$ and creating new models when required using the general memory $\mathscr{D}_G$. We adapt to \textit{real} drift by ensuring each model $M_i$ is the most updated mapping over its domain $\mathscr{M}_i$, with training augmented by our weak-supervision approach.

In our experimental work, we selected the normal distribution to model the sample density in any $\mathscr{M}_i$. Our next steps include incorporating kernel density estimation into $\rho$-band generation to allow higher degree of autonomy for our algorithm drift detection algorithm. We also plan to study further the drift in our data to identify any relations between drift types and errors propagated in neural and deep networks. This could aid in building more drift-resilient classifiers that are performant for longer periods of time in drifting conditions.

\bibliographystyle{aaai}
\bibliography{main}

\end{document}